\documentclass[10pt,twocolumn,letterpaper]{article}

\usepackage{cvpr}              

\usepackage[dvipsnames]{xcolor}

\definecolor{cvprblue}{rgb}{0.21,0.49,0.74}
\usepackage[pagebackref,breaklinks,colorlinks,citecolor=cvprblue]{hyperref}
\usepackage{algorithm}
\usepackage{algpseudocode}

\def\paperID{7} 
\def\confName{CVPR}
\def\confYear{2024}

\title{An Analysis on Quantizing Diffusion Transformers}

\author{Yuewei Yang, Jialiang Wang, Xiaoliang Dai, Peizhao Zhang, Hongbo Zhang\\
Meta GenAI\\
{\tt\small yueweiyang,jialiangw,xiaoliangdai,stzpz,hbzhang@meta.com}
}

\begin{document}
\maketitle
\begin{abstract}
    Diffusion Models (DMs) utilize an iterative denoising process to transform random noise into synthetic data. Initally proposed with a UNet structure, DMs excel at producing images that are virtually indistinguishable with or without conditioned text prompts. Later transformer-only structure is composed with DMs to achieve better performance. Though Latent Diffusion Models (LDMs) reduce the computational requirement by denoising in a latent space, it is extremely expensive to inference images for any operating devices due to the shear volume of parameters and feature sizes. Post Training Quantization (PTQ) offers an immediate remedy for a smaller storage size and more memory-efficient computation during inferencing. Prior works address PTQ of DMs on UNet structures have addressed the challenges in calibrating parameters for both activations and weights via moderate optimization. In this work, we pioneer an efficient PTQ on transformer-only structure without any optimization. By analysing challenges in quantizing activations and weights for diffusion transformers, we propose a single-step sampling calibration on activations and adapt group-wise quantization on weights for low-bit quantization. We demonstrate the efficiency and effectiveness of proposed methods with preliminary experiments on conditional image generation.
\end{abstract}
\section{Introduction}
Diffusion Models (DMs) \cite{ho2020denoising,sohl2015deep} represent a powerful class of generative models that have gained significant attention in recent years due to their ability to generate high-quality synthetic data. These models operate by iteratively denoising random noise to generate synthetic data, offering promising applications in various fields, including computer vision \cite{liu2023diffusion,chen2023diffusiondet,han2022card}, natural language processing \cite{lovelace2024latent,li2022diffusion,wu2024ar}, and image generation \cite{nichol2021glide,dhariwal2021diffusion,rombach2022high}. The UNet \cite{ronneberger2015u} architecture is a popular variant of diffusion models that incorporates a U-shaped network topology. This architecture is characterized by a contracting path, which captures coarse-grained features, and an expansive path, which facilitates the reconstruction of detailed structures. Dms with UNet structure have demonstrated efficacy in various image generation tasks \cite{saharia2022photorealistic,zhang2023adding,ruiz2023dreambooth}, offering a balance between efficiency and performance \cite{yang2023diffusion,croitoru2023diffusion}. In contrast, the transformer-only structure eschews convolutional layers in favor of a transformer architecture, which excels in capturing long-range dependencies and global context. Transformer-only diffusion models \cite{peebles2023scalable,chen2023pixartalpha} are a versatile and potent approach to generative modeling, boasting effective long-range dependency modeling, parallelizable computation, adaptability across data modalities and resolutions, alongside inherent interpretability. Nonetheless, the computational demand is inherently high because of the extensive storage space needed for parameters and the opulent memory required for inferencing features. Post-Training Quantization (PTQ) mitigates this demand by compressing model parameters after training, thereby reducing computational operations during inference. Previous studies \cite{li2023q,he2024ptqd,shang2023post} have tackled the challenge of quantizing UNet-structured DMs concerning both activations and weights. While some calibration techniques can be extended to diffusion transformers, these approaches necessitate moderate optimization efforts and rely on UNet-specific structures, such as shortcut connections. In this research, we pioneer the investigation of quantizing a transformer-only diffusion model without any optimizations. By addressing the activation quantization challenge through calibration with single-step sampling and tackling the weight quantization challenge through group-wise quantization adaptation, we showcase the effectiveness and efficiency of our proposed method in the text-to-image generation task.

\section{Related Work}
\noindent\textbf{PTQ for Diffusion Models} Quantizing DMs involves reducing numerical precision to enhance model efficiency and minimize operational size, with PTQ refining the model post-training using calibration data. While quantization shows promise for Large Language Models (LLMs) \cite{xiao2023smoothquant, bai2022towards}, adapting it to DMs presents unique challenges due to their iterative and dynamic denoising steps, exacerbating quantization of activations containing dynamic outliers through simple linear quantization. Previous works \cite{li2023q, he2023ptqd, wang2023towards} address these challenges by selecting uniformly distributed calibration data across inference timesteps and optimizing quantization parameters for dynamic activation ranges. While all aforementioned works focusing on UNet-structured DMs, in this paper, we present an analysis focusing on transformer-only DMs to propose an efficient and effective quantization strategy without any optimizations.

\noindent\textbf{Quantization of Diffusion Transformers} To the best of our knowledge, no prior research has specifically explored the quantization of diffusion transformers. However, existing studies have investigated challenges related to quantizing transformers for vision tasks. For instance, \cite{xu2023q} proposes a method that distills knowledge from the parent model to correct query information distortion. Similarly, \cite{li2023q,li2024bi} employ Quantization-Aware Training (QAT) to mitigate information distortion in self-attention maps. These approaches primarily target distortions in the attention mechanism and typically require substantial retraining efforts. Moreover, the dynamic activation ranges observed during multiple sampling steps exacerbate quantization errors cumulatively. In this study, we direct our attention toward addressing the challenge of quantizing dynamic activations and dispersed weights without the need for retraining, using Post-Training Quantization (PTQ).

\section{Method}

\begin{figure*}[t]
    \centering
    \includegraphics[scale=0.4]{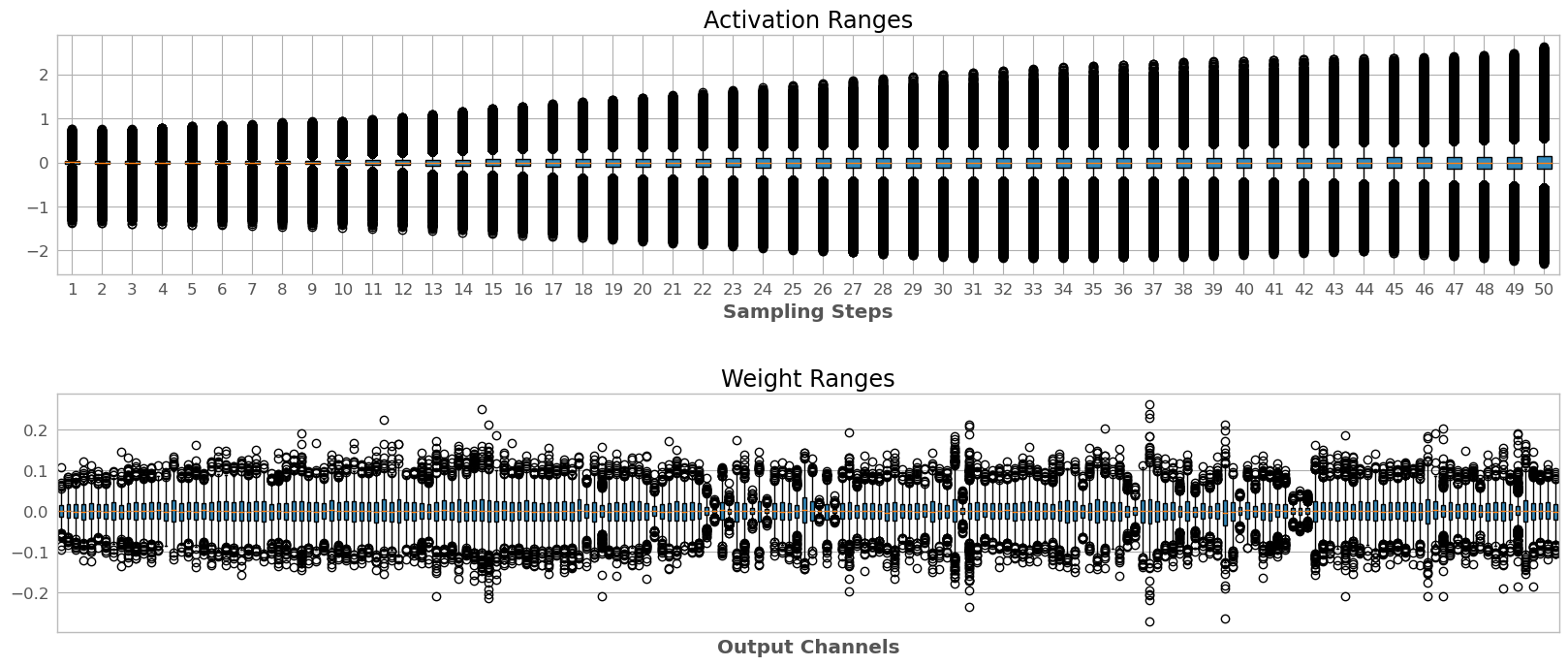}
    \caption{Without optimizations, data ranges pose challenges to quantize both activations and weights especially at a lower bit-width. (\textbf{Above}) Activation range varies dynamically across sampling steps and significant outliers persist. (\textbf{Below}) Weights are quantized channel-wise, but dispearsed outliers for each channel introduces high quantization loss when compressed to a lower bit.}
    \label{fig:act-weight-range}
\end{figure*}

\subsection{Post Training Quantization}
Post-training quantization (PTQ) reduces numerical representations by rounding elements $v$ to a discrete set of values, where the quantization and de-quantization can be formulated as:
\begin{equation} \label{eq:quantdequant}
    \hat{v} = s\cdot{clip}(round(v/s),c_{min},c_{max})
\end{equation} 
where $s$ denotes the quantization scale parameters. round(·) represents a rounding function \cite{cai2020zeroq,wu2020easyquant}. $c_{min}$ and $c_{max}$ are the lower and upper bounds for the clipping function $clip(\cdot)$. Calibrating parameters through weight and activation distribution estimation is crucial. The quantization error can be measured as the L2 norm of the Euclidean distance between quantized and unquantized elements:
\begin{equation*}
    Error = ||\hat{v}-v||_2^2
\end{equation*}

\subsection{Diffusion Models}
DMs \cite{sohl2015deep,ho2020denoising} involve two processes: forward diffusion and reverse diffusion. Forward diffusion process adds noise to input image, $\mathbf{x}_0$ to $\mathbf{x}_T$ iteratively and reverse diffusion process denoises the corrupted data, $\mathbf{x}_T$ to $\mathbf{x}_0$ iteratively. 
The forward process adds a Gaussian noise, $\mathcal{N}(\mathbf{x}_{t-1};\sqrt{1-\beta_t}\mathbf{x}_{t-1},\beta_t \mathbb{I})$, to the example at the previous time step, $\mathbf{x}_{t-1}$. The noise-adding process is controlled by a noise scheduler, $\beta_t$.
The reverse process aims to learn a model to align the data distributions between denoised examples and uncorrupted examples at time step $t-1$ with the knowledge of corrupted examples at time step $t$. 
To simplify the optimization, \cite{ho2020denoising} proposes only approximate the mean noise, $\theta(\mathbf{x}_{t},t)\sim \mathcal{N}(\mathbf{x}_{t-1};\sqrt{1-\beta_t}\mathbf{x}_{t},\beta_t \mathbb{I})$, to be denoised at time step $t$ by assuming that the variance is fixed. So the reverse process or the inference process is modeled as:
\begin{align}
    \begin{split}
        unconditional:p_\theta(\mathbf{x}_{t-1} \vert \mathbf{x}_t) &= \theta(\mathbf{x}_{t},t) \\
        conditional:p_\theta(\mathbf{x}_{t-1} \vert \mathbf{x}_t) &= \theta(\mathbf{x}_{t},t,\tau(y))
    \end{split}
\end{align}
where $y$ is a conditional context (i.e. class labels or text prompts).

Initially, UNet was a popular choice for $\theta$, but later studies \cite{peebles2023scalable,chen2023pixartalpha} replaced it with a transformer-only structure. Diffusion transformers differ from UNet models by having fewer convolutional layers, heavily utilizing linear layers, and lacking shortcut connections between downsampling and upsampling stages. Because of these differences, prior research on quantizing diffusion UNet may not adapt well to diffusion transformers, particularly when quantizing weights to lower bits.

\subsection{PTQ for Diffusion Transformers}
We identify challenges in quantizing activations in a diffusion process and quantizing weights in a transformer-only diffusion model by examine the data ranges for activations across sampling steps and weights across individual output channels for a linear layer in a diffusion transformer. 

\subsubsection{Calibrate Activation with 1-step Calibration}
\begin{figure}[htbp]
    \centering
    \includegraphics[scale=0.35]{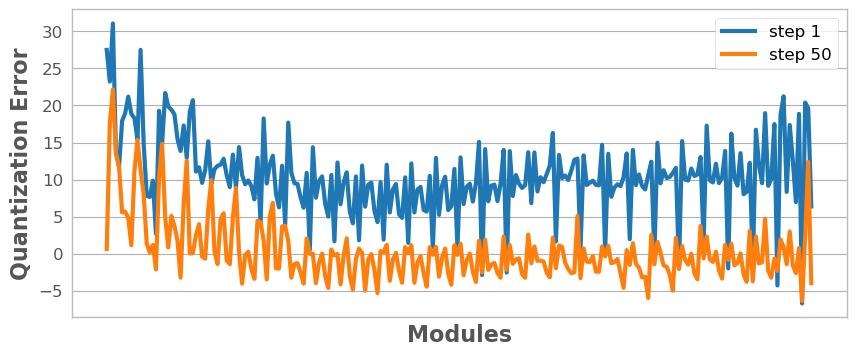}
    \caption{The quantization error is measured in $\mathbf{SQNR}$. There is a discrepancy between the first and the last sampling process. The calibrated paramters are most robust when the added noise is the strongest.}
    \label{fig:quant_error}
\end{figure}

\begin{figure}[htbp]
     \centering
     \begin{subfigure}[b]{0.13\textwidth}
         \centering
         \includegraphics[width=\textwidth]{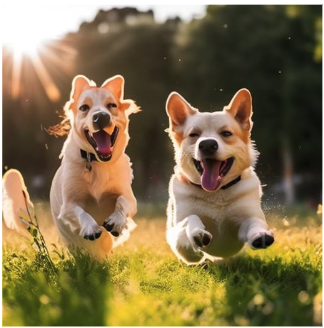}
         \caption{Full Precision}
         \label{fig:fp}
     \end{subfigure}
     \hfill
     \begin{subfigure}[b]{0.13\textwidth}
         \centering
         \includegraphics[width=\textwidth]{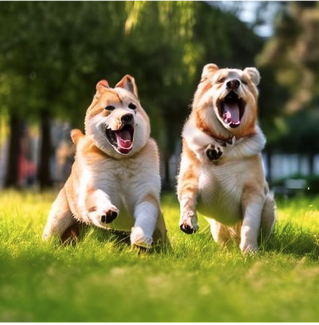}
         \caption{1-step}
         \label{fig:1step}
     \end{subfigure}
     \hfill
     \begin{subfigure}[b]{0.13\textwidth}
         \centering
         \includegraphics[width=\textwidth]{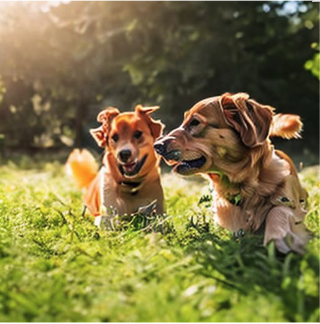}
         \caption{50-step}
         \label{fig:50steps}
     \end{subfigure}
        \caption{Calibrating through 50 steps produces visible image noise. 1-step calibration generates quality closer to the full precision output. Outputs from conditional DiT 8A8W.}
        \label{fig:sampling}
\end{figure}

Fig \ref{fig:act-weight-range} highlights the substantial variation in activations across sampling steps, a common issue also acknowledged in prior UNet DM quantization studies. Without optimization, calibrated parameters ($c_{min}$ and $c_{max}$) display significant fluctuation between initial and final sampling steps, leading to inconsistent stability throughout different stages. This phenomenon results from incrementally introduced noise during forward diffusion, which triggers estimated noise to adapt to gradual changes. We assess the robustness of DiT parameterized components against an 8A4W low bit quantization setting (8-bit activations and 4-bit weights). Fig \ref{fig:quant_error} displays the $\mathbf{SQNR}$ (\ref{eq:sqnr}) based quantization error between unquantized and quantized features, showing better resilience for calibrated parameters at the initial reverse diffusion step where noise is highest. Thus, we recommend using a single sampling step with the noise scheduler for quantization parameter calibration due to less varying activations and maximized noise, leading to more robust parameters overall. See Fig \ref{fig:sampling} for examples contrasting 1-step and 50-step calibrations.

\begin{equation}\label{eq:sqnr}
    \mathbf{SQNR}_{\theta,t} = 10log\mathbb{E}_{\mathbf{x}}\frac{||\theta_{fp}(\mathbf{x}_t)||^2_2}{||\theta_{q}(\mathbf{\hat{x}}_t)-\theta_{fp}(\mathbf{x}_t)||^2_2}
\end{equation}
where $\theta_{fp}$ is the full-precision model and $\theta_q$ is the quantized model, $\mathbf{\hat{x}}$ and $\mathbf{x}$ are quantized and unquantized output features.

\subsubsection{Calibrate Weights with Group Quantization}
In Fig \ref{fig:act-weight-range}, we also examine the ranges for weights of a linear layer in all output channels. Although weight quantization is performed per channel, many channels depicted in Figure \ref{fig:act-weight-range} display a substantial number of outliers scattered across them. This spread creates difficulties when attempting to represent all these weights using a reduced bit width. Indeed, directly applying 4-bit weights leads to severely distorted image outputs. Prior works have addressed this problem with QAT\cite{xu2023q} and finetuning\cite{liu2023pd}, but significant retraining and amount of data are required. Since we identified the dispersion of weight as the root cause of the problem, we propose to remedy the quantization difficulty with group-wise quantization. Previous works demonstrate that group-wise quantization is adaptable to transformers and is valid for hardware \cite{shen2020q,dettmers2023case}. In each channel, the weights can be further divided into groups. Each group will be calibrated individually. In this manner, the dispersed weights are divided into smaller ranges and therefore reduce the quantization difficulty for each group. It is simple to implement this group-wise quantization with just two lines. See Algorithm \ref{alg:group_quant} for detailed implementation.

\begin{algorithm}[htbp]
\caption{Group-wise Quantization}\label{alg:group_quant}
\begin{algorithmic}
\Require Weight:$\mathbf{w}\in\mathcal{R}^{C_{out}\times C_{in}}$, Group Size: $g$.
\Ensure $C_{in} \% g = 0$
\If{$g<C_{in}$}
    \State $shape=\mathbf{w}.shape$
    \State $\mathbf{w}=\mathbf{w}.reshape(-1,g)$
\EndIf
\State \textit{Calibrate}($\mathbf{w}$)
\If{$g<C_{in}$}
    \State $\mathbf{w}=\mathbf{w}.reshape(shape)$
\EndIf
\end{algorithmic}
\end{algorithm}
\section{Experiments and Results}

\begin{figure*}[htbp]
     \centering
     \begin{subfigure}[b]{0.4\textwidth}
         \centering
         \includegraphics[width=\textwidth]{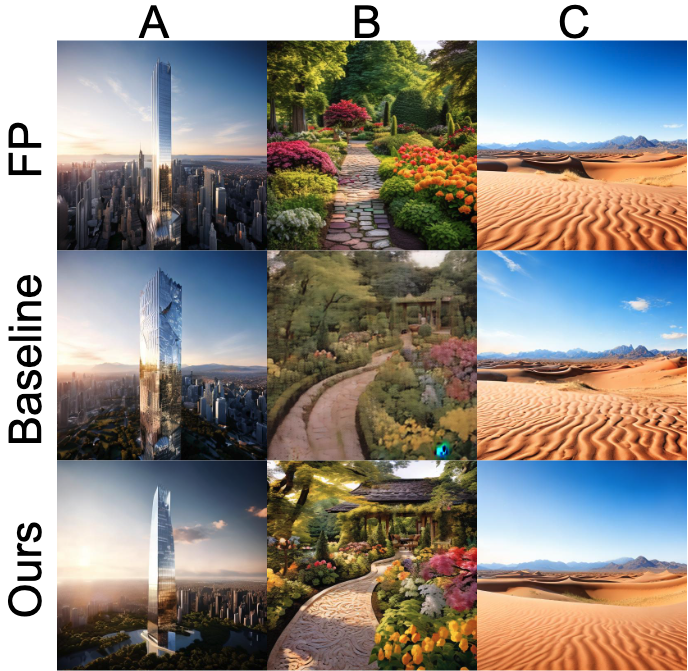}
         \caption{8A8W}
         \label{fig:8A8W}
     \end{subfigure}
     \begin{subfigure}[b]{0.4\textwidth}
         \centering
         \includegraphics[width=\textwidth]{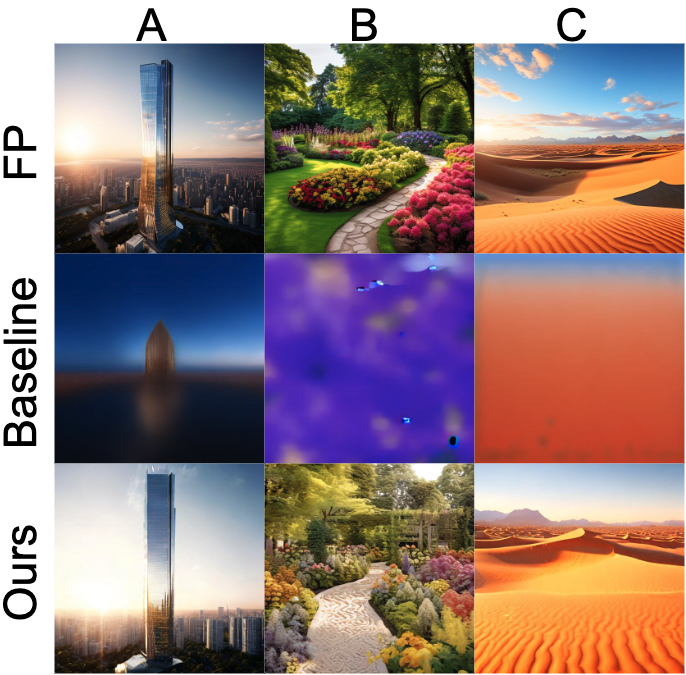}
         \caption{8A4W}
         \label{fig:8A4W}
     \end{subfigure}
    \caption{Qualitative examples of two quantization settings: 8A8W and 8A4W. Proposed improvements restore the original image quality.}
        \label{fig:examples}
\end{figure*}

\noindent\textbf{Dataset and Quantization Setting} We evaluate our quantization techniques for text-to-image generation under conditional settings using 1k captions. The evaluation set contains 30k randomly chosen captions from MS-COCO validation \cite{lin2014microsoft} dataset generations following prior methodologies \cite{gu2022vector,cao2018hashgan}. Classifier-free guidance is fixed at 3.0, and quantization is only applied to parameterized components (e.g., convolutional and fully connected layers). The default group size is 128 as in \cite{shen2020q}.

\noindent\textbf{Evaluation Metrics} For each experiment, two key metrics are provided: FID \cite{heusel2017gans} and $\mathbf{SQNR}_{\theta}$ averaged over sampling steps at the output to assess the generation of 512x512 images using Pixart Alpha. In addition to FID to COCO, we also include FID to full-precision model.

\noindent\textbf{Diffusion Setting}
We experiment on Pixart Alpha and with the sampling configuration listed in Table \ref{tbl:set}.

\begin{table}[htbp]
\centering
\resizebox{\linewidth}{!}{
\begin{tabular}{|l|l|l|}
\hline
Image Size & Inference Steps & Sampler                \\ \hline
512, 512   & 50              & DPMSolverMultistepScheduler          \\ \hline
\end{tabular}}
\caption{Sampling configuration.}
\label{tbl:set}
\end{table}

\begin{table}[htbp]
\resizebox{\linewidth}{!}{
\begin{tabular}{cccccc}
\hline
Bits(A/W)            & FID to COCO$\downarrow$ & FID to FP$\downarrow$ & $\mathbf{SQNR}_{\theta}$(db)$\uparrow$ \\ \hline
32fp/32fp            & 23.88      & 0.00      & -                                                 \\
8/8                  & 23.86    & 7.75       & 17.78                                             \\
+ 1 step calibration & 24.32 & 2.60  &  18.65  \\ 
8/4                  & 225.09    & 209.81       & 4.15                                             \\
+ 1 step calibration & 215.59 & 199.56  &  4.28     \\
+ group quantization & 22.13 & 5.16  &  12.36    \\ 
\hline
\end{tabular}}
\caption{1 step calibration significantly improves quality of generated images while maintaining a small size and fewer operations.}
\label{tbl:results}
\end{table}

\subsection{Discussions}
Table \ref{tbl:results} shows that our proposed method significantly surpasses baselines in both FID to FP and $\mathbf{SQNR}_{\theta}$ improvements. While FID is commonly used to assess generative models, its reliability has been doubted due to potential biases and limitations. Since the COCO datasets are photo-realistic and generated images diverge from this distribution, FID to COCO differences do not convey meaningful difference. Additional examples in Fig \ref{fig:examples} reveal that the enhanced models produce more realistic images with reduced quantization noise compared to baseline models.

\section{Conclusion}
In conclusion, this work represents a pioneering effort in exploring the quantization of transformer-only diffusion models without relying on any optimizations. By addressing activation quantization challenges via 1-step sampling calibration and overcoming weight quantization hurdles through group-wise adaptation, the proposed approach demonstrates both effectiveness and efficiency in the text-to-image generation task, providing valuable insights for further advancements in this area.
{
    \small
    \bibliographystyle{ieeenat_fullname}
    \bibliography{main}
}


\end{document}